# Planning Automated Driving with Accident Experience Referencing and Common-sense Inferencing

**Shaobo Qiu**[1], **Ji Li**[2], **Guoxi Chen**[3], **Hong Wang**[4], and **Boqi Li**[5]

[1, 2] Independent Researcher, CA, USA
{qiushaobo, hamletj }@gmail.com
[3] Dalian University of Technology, Dalian, P. R. China, chenguoxi@mail.dlut.edu.cn
[4] Tsinghua University, Beijing, P. R. China, hong_wang@mail.tsinghua.edu.cn
[5] University of Michigan, MI, USA, boqili@umich.edu

**Abstract**

Although a typical autopilot system far surpasses humans in term of sensing accuracy, performance stability and response agility, such a system is still far behind humans in the wisdom of understanding an unfamiliar environment with creativity, adaptivity and resiliency.

Current AD brains are basically expert systems featuring logical computations, which resemble the thinking flow of a left brain working at tactical level. A right brain is needed to upgrade the safety of automated driving vehicle onto next generation by making intuitive strategical judgements that can supervise the tactical action planning.

In this work, we present the concept of an Automated Driving Strategical Brain (ADSB): a framework of a scene perception and scene safety evaluation system that works at a higher abstraction level, incorporating experience referencing, common-sense inferring and goal-and-value judging capabilities, to provide a contextual perspective for decision making within automated driving planning.

The ADSB brain architecture is made up of the Experience Referencing Engine (ERE), the Common-sense Referencing Engine (CIE) and the Goal and Value Keeper (GVK). 1,614,748 cases from FARS/CRSS database of NHTSA in the period 1975 to 2018 are used for the training of ERE model. The kernel of CIE is a trained model, COMET-BART by ATOMIC, which can be used to provide directional advice when tactical-level environmental perception conclusions are ambiguous; it can also use future scenario models to remind tactical-level decision systems to plan ahead of a perceived hazard scene. GVK can take in any additional expert-hand-written rules that are of qualitative nature.

Moreover, we believe that with good scalability, the ADSB approach provides a potential solution to the problem of long-tail corner cases encountered in the validation of a rule-based planning algorithm.

## 1. Introduction

The automotive industry is using four effective approaches to assure the operational safety of automated driving (AD) vehicles. Firstly, it is ensuring functional safety by addressing the possible hazards caused by the malfunctioning behaviors of electrical and electronic safety-related systems. Secondly, it is following the design guidelines set forth by the <u>S</u>afety <u>O</u>f <u>T</u>he <u>I</u>ntended <u>F</u>unctionality (SOTIF), which refers to the absence of unreasonable risk due to hazards resulting from functional insufficiencies of the intended functionality or by reasonably foreseeable misuse by human beings. It accounts for corner cases that may give rise to the dangers that do not result from any system failures. Thirdly, it is observing operational design domain definitions, clearly specifying the operating conditions under which a given driving automation system or feature thereof is specifically designed to function. Fourthly, it is adopting safer human-machine interface (HMI) design.

Regardless of any comprehensive safety design countermeasures that are taken, and how thoroughly an automated driving system design is verified and validated, its effectiveness is confined within the functional boundary defined by the developers.

Human drivers can intuitively avoid entering complex driving environments that they cannot cope with and can rely on intuition to anticipate a situation in advance. The "road sense" and "driving intuition" possessed by a human driver's mind are capabilities that are currently missing from the motion planning of an automated vehicle. Intuition uses a combination of experience referencing and common-sense inferring. The human brain's intuitive judgment uses an empirical comparison method that is entirely different from logic computing, which is widely used in contemporary AD vehicles; it can provide approximate but quick solutions to unfamiliar problems. Although a typical autopilot system far surpasses humans in term of sensing accuracy, performance stability and response agility, such a system is still far behind humans in the wisdom of understanding an unfamiliar environment with creativity, adaptivity and resiliency.

Information science has presented the relationships among **d**ata, **i**nformation, **k**nowledge, and **w**isdom in a DIKW pyramid [1]. Data are symbols that represent properties of objects, events and their environments. They are products of sensing. Relationships between data are revealed and analysed to reach information. *Information* is the collection of *data* into groups of certain meaning, which is used for understanding patterns. This leads to accumulating *knowledge,* which is considered as actionable information that can answer "how" questions.

*Wisdom* is the capability of choosing and applying the right knowledge; it answers the "why" question [2].

The wisdom hierarchy arrangement corresponds rather well with the space-time theory of consciousness proposed by Kaku in his book *The Future of the Brain* ([3], pp 49), where he splits consciousness into four classes characterized by plants, reptiles, mammals and humans. In an AD intelligence framework, Kaku's four-class model can manifest as feedback response, action planning, behavior planning and mission planning. The correspondences among DIKW wisdom hierarchy, the four-class consciousness model and AV intelligence frameworks are summarised in Table 1.

Table 1 - DIKW wisdom hierarchy, the four-class consciousness model and AV intelligence frameworks

| DIKW Wisdom Hierarchy by Information science | Four-class Consciousness Model by [3] | AD Planning Policy Hierarchy |
|---|---|---|
| **Data** : Observed symbols that represent properties of objects, events and their environments. | **Class 0_ Plants**: Physical sensing | **Feedback Model**: Streaming content generated by sensors such as cameras, LiDAR, radar, ultrasonics, GPS and dynamic HD maps. |
| **Information** : Revealing relationships in the data to find an answer to "who", "what", "when", "where", or "how many" questions. | **Class I_ Reptile**: Space consciousness | **Time-space Model**: Identifying measures in terms of object classification, object positioning, object kinematics, and infrastructure property detection; collision prediction; etc. |
| **Knowledge**: Actionable information being able to answer "how" questions. | **Class II_ Mammal**: Social relations consciousness | **Social Relation Model**: Object behavior prediction; maneuver planning. |
| **Wisdom**: Principles and values are determined and "why" question is answered. | **Class III_ Human**: Future consciousness | **Inference Future Model**: Solution creation; alternatives evaluation; driving environment envision on a scenario level. |

Prevailing AD controllers employ expert systems that are good at accomplishing designated tasks specified by the designers based on physical rules. However, they are unable to anticipate future situation developments, to imagine events that are not detected by sensors, to recall lessons learned from historical traffic accidents, and to cope with driving environments that have not been defined by the algorithm developers. Current AD systems are capable of predicting object behavior and planning motions, have social relations consciousness, and can answer "how"-type questions, with an intelligence level of Class II_Mammal defined by [3], which is equivalent to the Knowledge-Level in the DIKW pyramid. For an AD vehicle to be able to deal with unfamiliar environments autonomously, it must be embedded with goals and values, be adaptative to the unknown, have future awareness, and be able to answer "why"-type questions, i.e., be at the Wisdom-Level in the DIKW pyramid, which is equivalent to the intelligent level of Class III_Human defined by [3]. For this purpose, the capability of an AD perception system should be above and beyond the picture description limited in mere physical relations, it should have the ability to make social relation predictions in the context of a certain driving environment, that is, to describe and predict the driving scenarios.

As human brain research demonstrated [48], humankind uses two brains for a range of accomplishments. The left brain is known as the logical side of the brain, handling reading, writing, and calculations. The right brain processes information in an intuitive and simultaneous manner which is more visual and deals in images. Similarly, current AD brains are basically expert systems featuring logical computations, which resemble the working flow of a left brain. In order to upgrade the safety of AD onto next generation, a right brain is needed to make intuitive strategical judgements that supervising the tactical action planning. The purpose of this paper is to propose a roadmap to achieve an advanced AD planning framework (hereafter referred to as AD s̲trategic b̲rain, or ADSB), endowed with sufficient resiliency and creativity to cope with unfamiliar environments, which means simulating the future based on evaluating the past, by incorporating historical accident lessons and human driving common sense, to advance the abstraction level of the AD vehicle perception system from *physical relation* level to *social relation* level.

Structure of the paper: Section 2 is dedicated to related works. In Section 3, we describe the roadmap to realise an ADSB. The module developments of an ADSB are presented in Section 4. We conclude with a discussion of future work in Section 5.

## 2. Background and Related Work

**Automated Driving Planning.** According to the order of execution, the automatic driving control chain can be divided into four stacks including detection, motion prediction, motion planning and motion execution, in which motion planning plays a critical part in determining the advancement level of an AD vehicle. A rule-based deterministic approach and data-driven probabilistic approach (AI planning) are two main routes applied throughout the whole stacks. Traditional rule-based policies are increasingly being replaced by neuro network policies in motion planning to gain higher planning accuracies.

Planning policies relying on rule-based algorithm perform decently in controlling L2/L3 vehicles, but they have a hard time solving the long tail problem for the L4/L5 corner-cases. The main reason is that rule-based planning algorithms lack scalability. Compared with the rule-based algorithm, machine learning algorithms have better scalability for absorbing new cases, so they are better at solving the long tail corner case problem. However, the main problem encountered by the machine learning algorithm is that the learning results exhibit insufficient interpretability and differentiability, which means it is not easy to trace the root cause of a problem, and to upgrade a local algorithm within a function module. Industry leaders are endeavouring to achieve the goals of interpretability, differentiability, and scalability simultaneously.

In 2019 Waymo introduced the ChauffeurNet system [4], the first system to use imitation learning to learn human driving while interfering with human driving behavior using artificial synthesis to achieve more robust probabilistic planning. ChauffeurNet does not have interpretability.

Lyft applied machine learning to prediction in 2020 [5] and introduced the concept of Autonomy 2.0 in 2021 [6]. Lyft calls rule-based artificial planning version 1.0, and 2.0 is characterized by data-driven and machine learning planning, with the aim of breaking through the scalability bottleneck of 1.0. Low-end sensor information is mainly used for the machine learning training input of 2.0. An imitation learning model was trained to imitate expert behavior. Differentiability is achieved between stacks.

In 2021, Uber used neural motion planner, MP3, to generate probabilistic intermediate representations, a more successful application of planning neural networks [7] [8]. MP3 starts with low-end sensing of raw information for end-to-end planning; it still requires LiDAR information to be inputted. MP3 is still not yet in large-scale practical use.

The 2021 SafetyNet system of Woven Planet (a subsidiary of Toyota) combined, for the first-time, a rule-based deterministic algorithm and ML based probabilistic planning results, striking a balance and complementarity between the two approaches. SafetyNet uses precise and interpretable artificial rules as a filtering layer, enabling Artificial Intelligence (AI) planning that relies solely on imitation learning, reducing collisions by 95% [9].

Tesla's FSD is known to rely solely on visual sensing for perception. The choice of the pure vision sensing route is not based on established theoretical and practical foundations, but on a belief that automated vehicles, like human beings, can extract enough visual information to deliver driving safety without relying on LiDAR and radar. FSD uses transformer technology to label 2D surrounding images into a unified 4D vector space (3D+1D time) [10].

Current automated vehicle planning focuses on solving the "how" question in completing a certain tactical action, without answering the "why" question to consider the strategical effectiveness of this action. Within the framework of the four-level awareness model, current AD planning is equivalent to Class 2 consciousness.

**Strategical Planning by AI.** Strategical planning overarches a series of actions that take in the future to achieve a final goal. Strategical planning requires the right balance of thinking ahead while actioning in the instant moment. It's the perfect convergence of what the future will hold to what must be done right now in order to make that future possible. The rule-based expert systems widely adopted in automated vehicles nowadays are not designed to provide adaptable and creative answers, therefore no strategical planning can be accomplished by such narrow AI systems. Thus, a broad artificial intelligence system, which features knowledge transfer capabilities, can sufficiently incorporate world or prior knowledge and adapt to changing tasks, is needed to accommodate unfamiliar driving environments.

M. Minsky ([11], pp163), argues that the incompetency shown by a present-day intelligent machine in DOIng daily chores is due to three reasons: It does not have common-sense knowledge; it does not have explicit goals, and its solutions are not resourceful enough. Broad AI is still in its infancy, and the challenge exists to develop systems that can combine learning techniques with inference techniques such as extrapolation and analogy

[12]. Nevertheless, abstracting knowledge at higher levels for current narrow AI can also substantially improve its strategic planning capabilities, which can determine importance priorities. Google's AlphaGo Zero is a good example.

AlphaGo is configured with two brains: a Policy Network and a Value Network, being integrated by a Monte-Carlo tree search. The Policy Network is aimed at achieving the most effective results by tactical actions, while the value network simulates human strategical thinking to guarantee that the trends develop in a favorable direction by looking at a bigger picture. As a result, AlphaGo grounds its deep thinking for immediate action on an intuitive judgement for the future [13], an approach that is typical of the "human mind" when dealing with complex problems.

**Human Consciousness Model.** Michio Kaku classifies the development of the consciousness of species into four stages: plant consciousness, reptile consciousness, mammalian consciousness, and the highest level: human consciousness, which "creates a model of the world and then simulates it in time, by evaluating the past to simulate the future", and this "requires mediating and evaluating many feedback loops in order to make decision to achieve a goal"([3], pp 49).

It is believed in [3] that in order for an artificial mind to have the ability to anticipate future events, both in the physical and social world like a human does, it must have firm grasp of the laws of nature, causality, and common sense. Emotions may also be a key to consciousness because they can determine what is important at this moment ([3], pp 223, pp 230).

D. Gilbert highlights "the ability to imagine objects and episodes that do not exist in the realm of the real" as the greatest achievement of human consciousness ([14], pp 5). The most important thing the human brain does is act as an "anticipation machine"; it is "making the future".

Although a human brain only consumes about 20 watts, it can quickly recall many existing modular "stereotypes" in cognitive thinking, so that it can quickly make judgments "by default", such as identifying objects, recognizing intentions, etc., with a very limited amount of computing. For this purpose, intuition that matches new information with experience/common sense is used to pull up the answers that are already there to quickly tackle unfamiliar problems. R. Gregory believes that perceptions are 90% (or more) stored knowledge, which is consistent with a brain anatomy discovery which shows that some 80% of fibers to the lateral geniculate nucleus relay station come downwards from the cortex, and only about 20% from the retinas [15].

Human being thinking was modelled by the combining two brains in ([11] pp 101): a tactical brain that gets signals from the external world and reacts but has no sense of what they might mean; a strategical brain that reacts to the decisions made by the tactical brain. The strategical brain cannot directly perform any physical actions, but it affects the external world by controlling the ways in which the tactical brain might react.

The strategical brain employs common-sense reasoning to decide what not to think about ([11] pp 140, pp 87, pp 145). We human don't perceive most fine details of things in the world until some parts of our minds make requests for them ([11] pp 158). To retrieve relevant knowledge with high efficiency, human brains have many ways to represent things. A few structures that our brains might use to represent the knowledge stored in our memories include stories or scripts, semantic networks, trans-frames, and knowledge-lines. ([11] pp 278).

Another factor supporting common-sense reasoning is the hierarchical organization of these representations. When our tactical systems start to fail, we humans are likely to engage our higher-level abstract ways of thinking. This is simpler and faster, because it suppresses the details that are not relevant. For the higher-level analogy to happen, one subject should present at least on three levels [11]: (1) descriptions of features, (2) descriptions of objects, and (3) descriptions of object relationships.

Most computer programs can still do only one particular kind of task, because they use only a single kind of representation and do only low-level abstraction. Higher layer abstraction and common-sense knowledge are indispensable to the construction of an artificial strategical brain.

**Machine Common Sense.** AI systems with common sense can make assumptions or default assumptions about the unknown similar to the way human being does. Common-sense knowledge also helps to solve problems confronting with incomplete information. The first AI program to address common sense knowledge was proposed by J. McCarthy.[16].

Machine common sense has long been a missing part of artificial intelligence (AI). Machine reasoning is narrow and has become highly specialized in recent advances in machine learning. Developers must carefully train or program systems for any situation based on physical rules. General common-sense reasoning remains elusive. The absence of common sense prevents intelligent systems from understanding their surrounding world, behaving reasonably in unforeseen situations, communicating naturally with people and learning from experiences.

A common-sense knowledge base has been a persistent challenge in AI research. From early efforts like CYC

[17] and WordNet [18], significant advances were achieved via the crowdsourced OpenMind Common Sense project, which lead to the crowdsourced ConceptNet [19] knowledge base. ATOMIC 2020, a new common-sense, general-purpose contains knowledge base that is not readily available in pretrained language models [20].

General AI with common sense is not yet available, but common-sense reasoning has successfully found applications in more limited domains such as natural language processing [21][22] and automated diagnosis [23] or analysis [24]. ConceptNet has also been used by chatbots [25] and by computers that compose original fiction.[26]. Consequently, applying common-sense knowledge in a single AI domain of AD is feasible.

## 3. The Construction of an AD Strategical Brain (ADSB)

This section sets up the guidelines to create an ADSB.

Before making any action decision, a humankind driver usually uses the following process to intuitively perceive the driving environment:

1) Building a rapid world model represented by highly simplified abstractions which contain the only information about identifications and about the relationship between the relevant surroundings
2) Retrieving experiences such as lessons learned in driving school, rules of thumb, or personal safety experience, to evaluate the safety of the current driving situation by making analogous comparisons between the reality and the past
3) Inferring the development of the current situation by using common sense to take advance precautious action in order to prevent a disadvantageous situation from evolving.

An ADSB will imitate the above process.

**AD Strategical Brain Motivation**. Due to the rule-based nature of algorithms, an AD tactical brain is unable to consider the scenario consequences of an action plan. A strategical brain, on the other hand, should possess the following capabilities:

1) Evaluates the safety of the current scene by referring to historical experience
2) Infers future scene development using common sense
3) Uses common sense to supplement the insufficiency of sensing capability.

Although the tactical brain has the advantages of accurate measurement, quick response, accurate action, no fatigue, and making good quantitative reasoning, the strategic brain is known for considering the consequences, focusing on the outcome, and making quick qualitative reasoning. The two kinds of brains have completely different missions, so the strategic brain is not an enhancement or supplement to the existing tactical brain. The tactical brain works on object behavior, while the strategical brain works on object relationship.

An AD dual-brain structure incorporating the wisdom features of human brain is envisioned as Fig. 1.

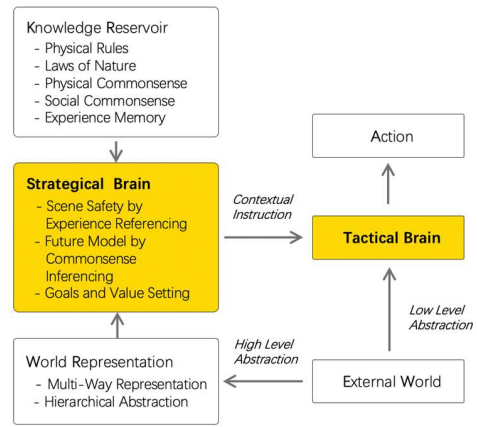

Fig. 1, Strategic Brain Tells the Tactical Brain How to React

**Abstraction Levels for World Representation of ADSB**. Current on-board computing power of AD systems have reached an astonishing level, but perception and decision-making operations are still limited to the tactical layer focusing on the present moment, that is, by observing the types, states and behaviors of objects. The bottom-line of ADSB function is the ability of envisioning a future scenario. There can be no strategic thinking without high-level abstractions ( [11] pp 64, pp 69). A strategical brain means carrying out thinking at higher semantic levels. Higher abstract descriptions are simpler because they suppress details that are not relevant. The good choice of a working abstraction involves removing as much detail as possible while retaining validity and ensuring that the abstract thinking is easier to carry out.

Perception levels are classified by [3] (pp 155) into low level, intermediate level, and higher semantic level, corresponding to feature cognition, object cognition, and scene cognition, respectively. ADSB's abstraction level is pinpointed at scene level.

Each research project and publication use its own definitions for terms such as situation, scene and scenario [27]-[32], [46], [47]. We find the definitions presented by [32] for such terms the most relevant to our purpose of sharing scene abstraction information across the experience reference, common-sense inference and scene perception platforms in this paper. More importantly, they can be used in a way agreeing on the content of existing road accident databases:

1) A *scene* describes a snapshot of the environment including the scenery and dynamic elements, as well as all actors' and observers' self-representations, and the relationships among those entities. Only a scene representation in a simulated world can be all-encompassing (objective scene, ground truth). In the real world it is incomplete, incorrect, uncertain, and from one or several observers' points of view (subjective scene).

2) A *scenario* describes the temporal development between several scenes in a sequence of scenes. Every scenario starts with an initial scene. Actions & events as well as goals & values may be specified to characterize this temporal development in a scenario. Other than a scene, a scenario spans a certain amount of time.

The relationship between a scene and a scenario is shown in Fig. 2 [32]. We eliminate the difference between situation and scene and discard the concept of a situation in [27].

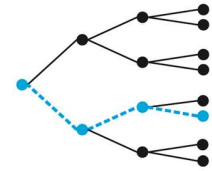

Fig. 2, A scenario (dashed blue) as a temporal sequence of actions/events (edges) and scenes (nodes)

Scenes are precisely what ADSBs are concerned about. Specifically, within the scope of this paper, a scene is further decomposed into three aspects of *scenery*, *actor* and the *relationships* between them. Aspects are further described by finer element. Element is the basic unit for ADSB algorithms to manipulate. *Scenery* is defined as the same as in [27], that is, the static aspects of the driving environment. The scene abstraction hierarchy for the purpose of ADSB is proposed in Fig. 3.

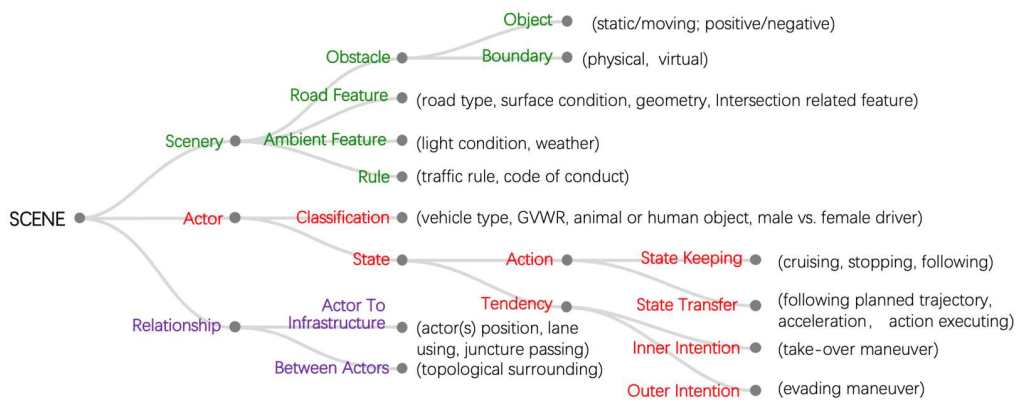

Fig. 3, ADSB Scene Abstraction Hierarchy

Accordingly, in AD systems, scenery contents can be sensed by the tactical brain, scenes can be judged by referencing experience, and scenarios have to be inferred by common sense. In this paper, *state* is defined as a maintained stable relationship with surrounding actors in an unchanged scenery; *event* means a transfer from one actor state to another state, and is usually the trigger of a scene.

**Experience Knowledge.** An accident "black spot" tells us the same as the rule of thumb that a driving school instructor teaches to learners: a typical accident pattern that repeatedly happened previously. The human brain recalls the experiences in order to make a quick analogous judgement and decision. Human driving experiences are memorized and represented in the abstraction level of scene description. We want to know from the experience memories the kinds of scene that are dangerous, the consequences a scene may result in, and the type of event that can trigger such a scene. Most of the current automated driving systems feature rule-based perception and planning at physical object level that are unable to take accident experiences into consideration.

Human beings can only observe scenes to envision potential danger; they are not good at accurate measurements. Nonetheless, they can still obtain approximate solutions for an unfamiliar situation by relying on very limited experiences. Safety experiences exist in the form of rule-of-thumb by word of mouth, driving school doctrines, traffic rules, road accident records, etc.

The public road accident database is a precious resource for an ADSB to reference. The most comprehensive data so far comes from the FARS (Fatality Analysis Reporting System) and the Crash Report Sampling System (CRSS), which are administered by the National Center for Statistics and Analysis (NCSA) within the NHTSA (National Highway Traffic Safety Administration) of the United States. The systems became operational in 1975 and contain data on a census of fatal traffic crashes in the 50 States, the District of Columbia, and Puerto Rico. To be included in FARS, a crash must involve a motor vehicle traveling on a trafficway customarily open to the public and must result in the death of an occupant of a vehicle or a non-occupant within 30 days (720 hours) of the crash [33],[34],[35]. FARS data are made available to the public in statistical analysis system (SAS) data files as well as comma-separated values (CSV) files. Unfortunately, no AD algorithm so far has taken advantage of this kind of resource.

For the current data collection year, there are 30 data files divided into CRASH, VEHICLE, PERSON, and EVENT levels, including more than 430 elements (see Fig. 4).

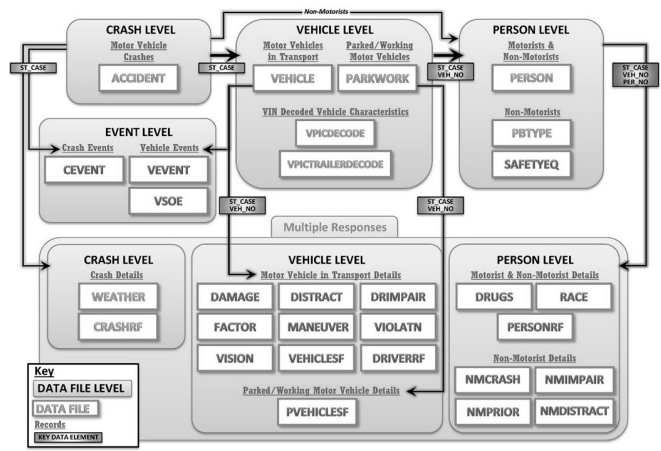

Fig. 4, FARS/CRSS Data File Levels [35]

FARS/CRSS data provide full support to describe the abstract level set out by Fig. 3, thus they are used by this paper to create the Experience Referencing Engine.

**Common-sense Knowledge Reservoir Requirement.**
Generally, common sense is partitioned into three dimensions [11], (i) Common sense of objects in the environment, including properties, theories (such as physics), and associated emotions; (ii) Common sense of object relationships, including taxonomic, spatial and structural relationships among the objects; (iii) Common sense of object interactions, including actions, processes, and procedural knowledge.

Common sense will be critically helpful to the ADSB motivations in the following ways: 1. inferring the scene development, 2. eliminating the ambiguity and increasing the interpretability of low-level abstractions by using common-sense-augmented perception technology.

Below are a few examples of exercising common-sense inferring; human drivers can answer the following questions or make anticipations:

1) In a continuous scene, if there are moving objects in the field of view that suddenly disappear or suddenly appear, is there a reasonable explanation for this topology discontinuity? Are there any occluded objects behind the visible objects?
2) Should the detected stationary object be a shadow or a real obstacle? Is it possible for this kind of obstacle to exist in this area? Is it possible for shadows to exist at this location and this time?
3) What factors will cause the transfer of the current scene? What kind of scene will it turn into?
4) Human drivers know that the same actor may exhibit totally different attributes or relationships in a different scene. For instance, a vehicle entering an "Exit Only" lane may suddenly return to the main lane and cut in, in front of you.
5) If a football suddenly flies into the lane, it will be very likely followed by a chasing boy rushing into the lane.
6) Driving in the "No-Zone" around a heavy vehicle should be avoided.
7) Give lorry drivers a free lane to enter the carriageway.
8) Keep a safety surrounding pattern to maintain an escape route in case there is a collision directly in front of the ego vehicle.

To this end, an ADSB common-sense reservoir should contain at least the following aspects of knowledge:

1) Know the kind of objects that may exist in the traffic environment
2) Know the properties of various objects
3) Know the space-time affiliation relationship between objects
4) Know the calibre of each actor's capability (lorries have more kinetic energy and are harder to accelerate and decelerate than sedans).
5) Know the causal interactions between actors (a sudden cut in will force the ego vehicle to brake or change lane)
6) Understand various actor reactions under a range of topological relationships between actors
7) Understand actor reactions under a range of sceneries (increase the following distance on icy roads)

8) The potential irregular behaviours of an actor situated in a particular scene
9) Interpretation of characteristic actor action (event chain inference)
10) Understand the implementation differences of traffic rules as scenery changes
11) Know the influence of dynamic scenery on perception results (shadow, occlusion, backlight), capable of making visual deductive human-like inferences
12) Be able to embody the rules of thumb

These competency questions are just a sketch and do not need to be exhaustive.

It should be pointed out that common sense is only used to understand scenery and foresee a scene chain. The responsibility of scene safety evaluation goes to experience analogous analysis carried out by the Experience Referencing Engine (see section 4 below).

In the driving safety domain, related objects and common-sense knowledge are numbered, and can edited using the ontology we chose below. This means the common-sense knowledge database can be programable and scalable, and it can be updated at any time without having to retrain the algorithm. We believe a compact common-sense knowledge base specific to the safety driving domain is achievable for the purpose of low computing power consumption and quick analysis. Rules of thumb, defensive driving doctrines, domain expert advice, and safety axioms are input by hand editing rather than by automatic crowdsourcing.

**Common-sense Knowledge Ontology**. The experience referencing, common-sense knowledge inferring and scene-perception platforms of ADSB are inter-operating agents; an ontology needs to be created for these platforms so that they can communicate within each other and draw upon knowledge from each other [36]. Ontologies are explicit formal specifications of the terms in the domain and relations among them [37]. Ontologies are used to establish concept agreements shared by experience referencing, common-sense inferring, and scene-sensing modules of ADSB, to integrate these varied domains into a coherent framework.

Ontology is a semantic tool inherently capable of inferring. To take full advantage of ontologies, reasoning has to be carried out. Several reasoners exist to achieve this task. These include Pellet, Fact++, Hermit and Racer [38]. They enable several inferencing supports, like fontology consistency checking, subsumption inference and class equivalence. [39].

Many researchers have utilized ontologies for advanced driver assistance systems or control of autonomous vehicles [29], [40], [41]. A simple ontology that includes context concepts such as actor entity (pedestrian and vehicle), static entity (road infrastructure and road intersection) and context parameters (isClose, isFollowing, and isToReach) is modelled in [41] to enable the vehicle to understand the context information when it approaches road intersections. By applying 14 rules written in Semantic Web Rule Language (SWRL), the ontology is capable of processing human-like reasoning on global road contexts.

Ontology is also indispensable in describing events and scenes. We use ontology to define the content of a scene hierarchy shown in Fig. 3. Ontology specified for a particular domain is easy to create [42], but, hopefully, it can be compatible with existing databases, such as the FARS/CRSS databases, and common-sense knowledge bases such as ConceptNet and WordNet.

Compared with ConceptNet and WordNet, we find the ontology of ATOMIC [20] to be more suitable for the purpose of application in this paper, because ATOMIC underlines the concept of events instead of objects (ConceptNet) and lexicons (WordNet).

This paper uses three kinds of nodes proposed by ATOMIC to describe driving scenes: event, social actor and physical entity, as well as 23 relations between nodes [46], as shown in Table 2, which supports the scene awareness hierarchy shown in Fig. 3 and the FARS/CRSS databases. By this structure, a node described by a sentence containing a subject, predicate and object elements is considered to be an *event*. "Event" means the completion of an actor *state* transfer, can further influence the development of a scene. Events can be described in the present continuous tense or in the subject-linkage-predicative structure, like:

*A Vehicle at Position_RR2 Is Changing into Its Left Line*

USDOT has devised the pre-crash scenes typology based primarily on pre-crash variables in the National Automotive Sampling System (NASS), crash databases including the General Estimates System (GES) crash database, and the Crashworthiness Data System (CDS) [43], which all depict vehicle movements and dynamics as well as the critical event occurring immediately prior to a crash [44]. Examples of relation aspect elements given in Fig. 5 agree on this typology.

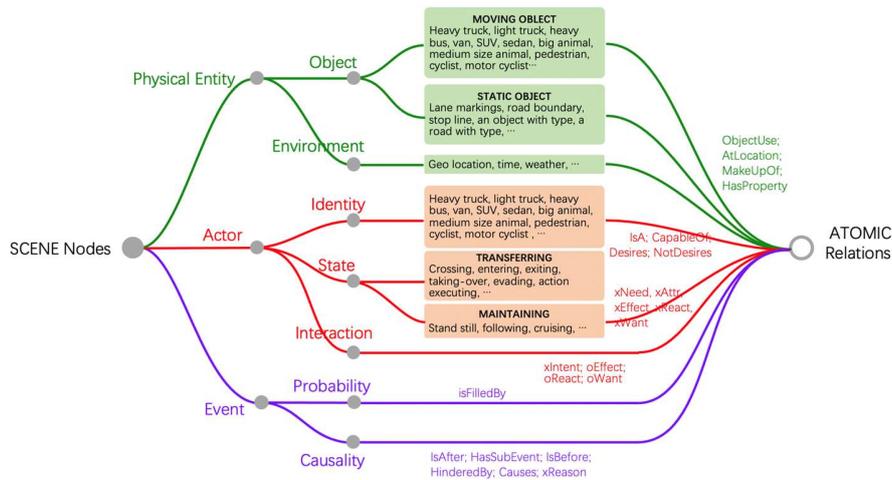

Fig. 5, Scene Node Types and the Relations Between them

**Goals and Values.** The human mind is considered an emotion machine by M. Minsky [11]. Emotions are essential—and not a luxury—for making decisions, because emotion decides what is important and what is not important. Goals and values determine the priority of investing resources. With goals and values at high-abstraction levels in mind, humans can choose priorities adaptive to the dynamic environment, and provide alternative solutions to unfamiliar questions in a resourceful way.

One major goal of ADSB is to maintain the vehicle drive in a *safety state*. We propose the definition of safety state in section 4, including Empirically Safe State, Theoretically Safe State, and Safe State of Rule-of-thumb.

**Scene Perception System.** For the purpose of evaluating a dynamic situation, the scene must be sensed in the first place. A scene is taken as a combination of a series of elements (Fig. 3), which have the same meaning as the element definition by NHTSA in section 2. Every element set out in 4.1 must be precepted to aggregate a current scene. In general, a scene perception system must be capable of:

1) Sensing dynamic scenery features
2) Classifying an object
3) Sensing the action of an actor
4) Sensing the topological relations among actors and infrastructures

**Difference Between Strategical Brain and Tactical Brain.** The strategical and tactical brains are responsible for qualitative reasoning and quantitative computing respectively. In its simplest form, qualitative reasoning is about the direction of change in interrelated quantities.

The designing of a strategical brain is not aimed at replacing or being an augmentation of the currently used tactical brain. The two brains work separately but complement each other. It is the combination of the two that differentiates humans from other animals. The strategical brain only qualitatively cares about the scene safety from the perspective of inter-relations, and whether there will be dangerous scenes in the future. It also answers "should", "would", "could" and "why" questions. It is not responsible for answering "what", "when", "where" and "how" questions. Such questions are answered quantitatively by the tactical brain using precise mathematical calculations. It has now reached a stage where the brain of the AD vehicle has to be split into two separate and complementary brains. In the framework of Kaku's four-level model [11], one example of interaction between the two kinds of AD brains is given in section 4.

The strategical brain allows for mistakes. This is partly because common-sense facts themselves have obscure exceptions, and also because accurate common-sense inferences can be of little relevance to particular driving situations. The strategical brain won't cause tactical brain crashes or interfere with the execution of a motion plan. This type of fail-soft is especially required when the tactical brain hesitates to make decision due to a lack of information.

A comparison between the strategical brain and the tactical brain is summarized in Table 2.

Table 2 - Differences Between Strategical Brain and Tactical Brain

|  | Tactical Brain | | Strategical Brain | |
| --- | --- | --- | --- | --- |
|  | In-Put | Out-Put | In-Put | Out-Put |
| Class 0:<br>Feedback Model | Object features | Feedback control | NA | NA |
| Class I:<br>Time-Space Model | Physical rules | Positioning; Classification; Relationship | Physical common-sense | Multi-layer abstraction; Scene description |
| Class II:<br>Social Relation Mosel | Individual behaviour paradigm | Intention estimation; Behaviour planning | Social behaviour common-sense | Interaction prediction |
| Class III:<br>Inferenced Future Model | NA | NA | Accident experiences; Defensive driving rules; Safety state goals | Safety state keeping; Solution synthesizing |

## 4. ADSB Module Developing

In this section we propose technical roadmaps for developing functional modules of ADSB.

**Experience Referencing Engine**. We trained an Experience Referencing Engine (ERE) model from the database provided by FARS/CRSS. The goal of ERE model training is to use this model to evaluate whether a currently observed scene is safe or not.

Elements of the FARS/CRSS database are used to aggregate a scene. A classification model was trained using the database, with scenes of each accident used as input, and the accident type and severity of each accident taken as output.

For the data collection year 2020, there are 30 data files divided into CRASH, VEHICLE, PERSON, and EVENT levels, including more than 430 elements. Not all of them are helpful for the purpose of ERE training; 81 elements among them are adopted by the training.

Annual changes are made to the type of data collected and the way data is presented in the data files. Some data files were discontinued at a certain year and new ones created. Coding rules have been changing. ID definitions for each data element and attributes for a given year have been changing. Coding manual provides a set of written instructions on how to transfer the information from a police crash report to the FARS data [33]. The manual presents the evolution of FARS coding from inception through to the present day [34] [35].

In FARS/CRSS, an accident is represented by *elements* which are described by many *attribute* values. Each element has a name and a locator ID. Element attribute values are determined by a coding system defined by the NHTSA [33]. For example, an element describing first harmful event is given a name HARM_EV, and is assigned an ID (locator code) C19. The attributes code of the element could have 65 values. Unfortunately, the meanings of IDs and code values could vary from year to year. For example, when its attribute code is 55 for element C19, that means "Other Not in Transport Motor Vehicle" for the years 2005–2007, but it otherwise means "Motor Vehicle in Motion Outside the Trafficway" since 2008. Another example: in 2022, the ID of element "Most Harmful Event" was changed from V33 to V38. Furthermore, there is a continuous discontinuing of old attributes or adding of new ones.

Before the start of algorithm training, a pre-processing unit that can automatically normalise the element ID changes and element attribute code changes is needed for the scenes of an accident to be retrieved correctly for the purpose of training. The pre-processing Unit in Fig. 6 accomplishes this and transforms raw FARS/CRSS data into a consolidated dataset trainable by the classification model of the prediction algorithm training module.

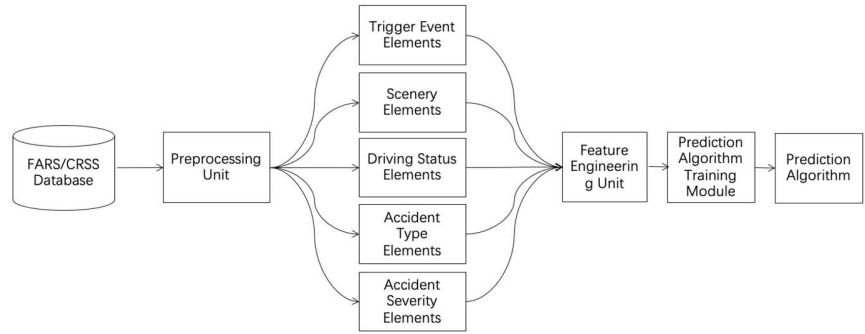

Fig. 6, Experience Referencing Engine (ERE) Training

Moreover, the pre-processing unit divides elements into two categories: causal elements and effect elements. Causal elements are close to what the human senses can perceive, including scenery elements, trigger event elements, and driving status elements. Effect elements include accident type and accident severity.

Selected causal elements are summarized in Table 3. Effect Elements are summarized in Table 4.

Table 3 - Causal Elements

| Input Module | Element Name (ID (Locator Code)) as per [3] |
|---|---|
| Scenery Elements | County(C6), City (C7), Month of Crash (C8A), Day of Crash (C8B), Day of Week (C8C), Year of Crash (C8D), Hour of Crash( C9A), Minute of Crash( C9B), Trafficway Identifier (C10), Route Signing (C11), Milepoint (C16), Latitude (C17A), Longitude (C17B), Relation to Junction (C21B), Type of Intersection (C21B), Relation to Trafficway (C23), Work Zone (C24), Light Condition (C25), Atmospheric Conditions (C26), School Bus Related (C27), Rail Grade Crossing Identifier (C28), Body Type (V11), Vehicle Trailing (V14), Gross Vehicle Weight Rating (V18), Vehicle Configuration (V19), Cargo Body Type (V20), Bus Use (V22), Special Use (V23), Emergency Use (V24), Truck Weight Rating (V123), Travel Speed (V25), Driver's ZIP Code (D6), Non-CDL License Type (D7A), Trafficway Description (PC5), Total Lanes in Roadway (PC6), Speed Limit (PC7), Roadway Alignment (PC8), Roadway Grade (PC9), Roadway Surface Type (PC10), Roadway Surface Condition (PC11), Driver's Vision Obscured by (PC14), Marked Crosswalk Present (NM9-PB27), Sidewalk Present (NM9-PB28), X.School Zone (NM9-PB29) |
| Trigger Event Elements | Critical Event Precrash (PC19), Related Factors- Crash Level (C32), Attempted Avoidance maneuver (PC20), Crash Type-Pedestrian (NM9-PB30), Pre-Event Movement (Prior to Recognition of Critical Event) (PC17), Marked Crosswalk Present (NM9-PB27), Sidewalk Present (NM9-PB28), School Zone (NM9-PB29), Crash Type -Pedestrian (NM9-PB30), Crash Type-Bicycle (NM9-PB30B), Crash Location-Pedestrian (NM9-PB31), Crash Location-Bicycle (NM9-PB31B), Pedestrian Position (NM9-PB32), Bicyclist Position (NM9-PB32B), Pedestrian Initial Direction of Travel (NM9-PB33), Bicyclist Initial Direction of Travel (NM9-PB33B), Motorist Initial Direction of Travel (NM9-PB34), Motorist maneuver (NM9-PB35), Intersection Leg (NM9-PB36), Pedestrian Scenario (NM9-PB37), Crash Group-Pedestrian (NM9-PB38), Crash Group-Cycle (NM9-PB38B) |
| Driving Status Elements | Related Factors: Driver Level (D24), Sex (P6/NM6), Age (P5/NM5) |

Table 4 - Effect Elements

| Output Module | Element Name (ID (Locator Code)) as per [3] |
|---|---|
| Accident Types | First Harmful Event (C19), Jack-knife (V22), Rollover (V32), |
| Accident Severity | Vehicle Number (V3/D3/PC3/P3/NM4 ), Extent of Damage (V35), Ejection (P13), Jack-knife (V22), Rollover (V32), Fire Occurrence(V39), Fatals (V150), Injury Severity (P8/NM8) |

Accident type are listed in Table 5 as defined in [35], and the illustrations of which can be found in Appendix A: PC23 Crash Type Diagram of [35]

Table 5 - Element Attributes of Element "Crash Type" (PC23)

| CATEGORY | CONFIGURATION | Attribute Codes and Element Attributes |
|---|---|---|
| I: SINGLE DRIVER | A: RIGHT ROADSIDE DEPARTURE | 1 Drive off Road, 2 Control/Traction Loss, 3 Avoid Collision With Vehicle, Pedestrian, Animal, 4 Specifics Other, 5 Specifics Unknown |
| | B: LEFT ROADSIDE DEPARTURE | 6 Drive off Road, 7 Control/Traction Loss, 8 Avoid Collision With Vehicle, Pedestrian, Animal, 9 Specifics Other, 10 Specifics Unknown |
| | C: FORWARD IMPACT | 11 Parked Vehicle, 12 Stationary Object, 13 Pedestrian/Animal, 14 End Departure, 15 Specifics Other, 16 Specifics Unknown |
| II: SAME TRAFFICWAY, SAME DIRECTION | D: REAR END | 20 Stopped, 21 Stopped, Straight, 22 Stopped, Left, 23 Stopped, Right, 24 Slower, 25 Slower, Going Straight, 26 Slower, Going Left, 27 Slower, Going Right, 28 Decelerating (Slowing), 29 Decelerating (Slowing), Going Straight, 30 Decelerating (Slowing), Going Left, 31 Decelerating (Slowing), Going Right, 32 Specifics Other, 33 Specifics Unknown |
| | E: FORWARD IMPACT | 34 Control/Traction Loss, Avoiding Non-Contact Vehicle- Vehicle's Frontal Area Impacts Another Vehicle, 35 Control/Traction Loss, Avoiding Non-Contact Vehicle- Vehicle Is Impacted by Frontal Area of Another Vehicle, 36 Control/Traction Loss, Avoiding Non-Fixed Object- Vehicle's Frontal Area Impacts Another Vehicle, 37 Control/Traction Loss, Avoiding Non-Fixed Object- Vehicle Is Impacted by Frontal Area of Another Vehicle, 38 Avoiding Non-Contact Vehicle- Vehicle's Frontal Area Impacts Another Vehicle, 39 Avoiding Non-Contact Vehicle- Vehicle Is Impacted by Frontal Area of Another Vehicle, 40 Avoiding Non-Fixed Object- Vehicle's Frontal Area Impacts Another Vehicle, 41 Avoiding Non-Fixed Object- Vehicle Is Impacted by Frontal Area of Another Vehicle, 42 Specifics Other, 43 Specifics Unknown |
| | F: SIDESWIPE/ANGLE | 44 Straight Ahead on Left, 45 Straight Ahead on Left/Right, 46 Changing Lanes to the Right, 47 Changing Lanes to the Left, 48 Specifics Other, 49 Specifics Unknown |
| III: SAME TRAFFICWAY, OPPOSITE DIRECTION | G: HEAD-ON | 50 Lateral Move (Left/Right), 51 Lateral Move (Going Straight), 52 Specifics Other, 53 Specifics Unknown |
| | H: FORWARD IMPACT | 54 Control/Traction Loss, Avoiding Non-Contact Vehicle- Vehicle's Frontal Area Impacts Another Vehicle 55 Control/Traction Loss, Avoiding Non-Contact Vehicle- Vehicle Is Impacted by Frontal Area of Another Vehicle, 56 Control/Traction Loss, Avoiding Non-Fixed Object- Vehicle's Frontal Area Impacts Another Vehicle, 57 Control/Traction Loss, Avoiding Non-Fixed Object- Vehicle Is Impacted by Frontal Area of Another Vehicle, 58 Avoiding Non-Contact Vehicle—Vehicle's Frontal Area Impacts Another Vehicle, 59 Avoiding Non-Contact Vehicle—Vehicle Is Impacted by Frontal Area of Another Vehicle, 60 Avoiding Non-Fixed Object- Vehicle's Frontal Area Impacts Another Vehicle, 61 Avoiding Non-Fixed Object- Vehicle Is Impacted by Frontal Area of Another Vehicle, 62 Specifics Other, 63 Specifics Unknown |
| | I: SIDESWIPE/ANGLE | 64 Lateral Move (Left/Right), 65 Lateral Move (Going Straight), 66 Specifics Other, 67 Specifics Unknown |
| IV: CHANGING TRAFFICWAY, VEHICLE TURNING | J: TURN ACROSS PATH | 68 Initial Opposite Directions (Left/Right), 69 Initial Opposite Directions (Going Straight), 70 Initial Same Directions (Turning Right), 71 Initial Same Directions (Going Straight), 72 Initial Same Directions (Turning Left), 73 Initial Same Directions (Going Straight), 74 Specifics Other 75 Specifics Unknown |
| | K: TURN INTO PATH | 76 Turn Into Same Direction (Turning Left), 77 Turn Into Same Direction (Going Straight), 78 Turn Into Same Direction (Turning Right), 79 Turn Into Same Direction (Going Straight) 80 Turn Into Opposite Directions (Turning Right), 81 Turn Into Opposite Directions (Going Straight), 82 Turn Into Opposite Directions (Turning Left), 83 Turn Into Opposite Directions (Going Straight), 84 Specifics Other, 85 Specifics Unknown |
| V: INTERSECTING PATHS (VEHICLE DAMAGE) | L: STRAIGHT PATHS | 86 Striking From the Right, 87 Struck on the Right, 88 Striking From the Left, 89 Struck on the Left, 90 Specifics Other, 91 Specifics Unknown |
| VI: MISCELLANEOUS | M: BACKING, ETC. | 92 Backing Vehicle, 93 Other Vehicle or Object (2010-2012), 93 Other Vehicle (2013–Later), 98 Other Crash Type, 99 Unknown Crash Type |

For the purpose of manageable complexity, a comprehensive assessment measure of both the degree of personal injury and the degree of damage to the vehicle or other property is introduced. the comprehensive severity index (CSI) is defined as:

CSI = 10a+6b+4c+3d+2e+2f+2g

in which, a, b, c, d, e, f, g is calculated by Table 6.

Severity is classified into five levels: Level I when CSI ≤ 2, Level II when CSI = 2–5, Level III when CSI = 6–9, Level IV when CSI = 10–14, Level V when CSI ≥ 15.

Table 6 - Coefficient Determination for CSI Calculation

| Elements | Attributes Included | Measure |
|---|---|---|
| Fatals (V150) | Reported Number | a = Number of fatalities |
| Injury Severity (P8/NM8) | 3, 4, 5 | b = The total number of persons with attribute 3, 4, 5 |
| Ejection (P13) | 1, 2, 3 | c = The total number of persons with attribute 1, 2, 3 |
| Extent of Damage (V35) | 6 | d = Number of vehicles with attribute 6 |
| Rollover (V32) | 1, 2, 9 | e = Number of vehicles with attribute 1, 2, 9 |
| Fire Occurrence(V39) | 1,2 | f = Number of vehicles with attribute 1, 2 |
| Jack-knife (V22) | 2, 3 | g = Number of vehicles with attribute 2, 3 |

Fig. 7 shows components of the prediction algorithm in Fig. 6. The prediction algorithm includes a similar accident search unit, a severe damage detection unit, and a severity rating unit. The similar accident search unit searches for similar historical accident records if the input scene is analogous to the attribute combination of a certain recorded accident. If a similar accident exists in the accident database, the severe damage detection unit determines whether it is likely that the accident will include severe damage. If that is likely, the severity rating unit rates the severity of the accident.

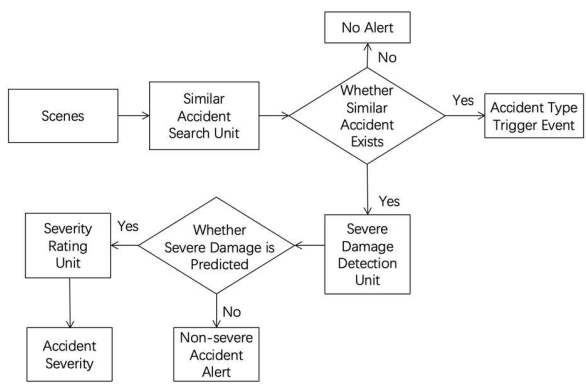

Fig. 7 Components of the Prediction Algorithm

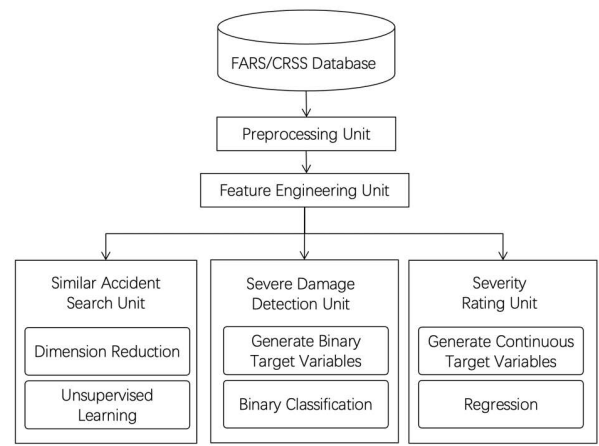

Fig. 8, Learning Model Training Flowchart of Prediction Module

Fig. 8 is a flowchart of machine learning model training of the Prediction Algorithm Training Module in Fig. 6. As described above, the pre-processing unit outputs a consolidated dataset suitable for machine learning training. The Feature Engineering Unit takes into consideration each of the fields in Table 3. The Feature Engineering Unit processes each field of Table 4, 5 by grouping information, creating new variables (such as finding the difference between travel speed and speed limit), etc. The produced data is then transmitted to each of the Similar Accident Search Unit, the Severe Damage Detection Unit, and the Severity Rating Unit.

In machine learning model training, the Similar Accident Search Unit is an unsupervised learning model. More specifically, a clustering model, using K-means and

HDBSCAN. Before clustering, dimension reduction is used to reduce the dimension of the information.

The severity score is estimated by CSI, according to the attributes in Table 6. The severity rating is then derived from CSI by bucketing severity score into 5 levels. The binary severity, severe: 1 or non-severe: 0, is the derived by mapping the severity level 1 to non-severe 0, and mapping the severity level larger than 1 to 1. A positive value means a severe accident, and a negative value means a non-severe accident.

The Severe Damage Detection Unit is a binary classification model with the above binary severities as target variables. This model is developed using a Random Forest Classifier. It detects either there will be a severe damage or not. The Severity Rating Unit is a step beyond the Severe Damage Detection Unit. It is a multi-class classification model with the above 5 level severities as target variables. It is developed using a Random Forest Classifier. It predicts the severity level for detected future severe damage.

In machine learning model training, the similar accident search unit is an unsupervised learning model. More specifically, it is a clustering model, using K-means and HDBSCAN. Before clustering, dimension reduction is used to reduce the dimension of the information, because typically the dimension is too large to obtain good performance.

There are 1,614,748 cases from FARS and CRSS in the period 1975 to 2018 being used for the model building, where 80% are used for training and 20% remain untouched for algorithm validation, examples of which are shown in Fig. 9 and Fig. 10.

```
              precision    recall  f1-score   support

           0       0.82      0.77      0.80     65344
           1       0.94      0.96      0.95    257606

    accuracy                           0.92    322950
   macro avg       0.88      0.86      0.87    322950
weighted avg       0.92      0.92      0.92    322950
```

Fig. 9, Severe Damage and Accident Type Detection Report

```
              precision    recall  f1-score   support

           1       0.78      0.79      0.78     65344
           2       0.46      0.53      0.49     77547
           3       0.45      0.52      0.49     78375
           4       0.36      0.30      0.33     48866
           5       0.43      0.29      0.34     52818

    accuracy                           0.51    322950
   macro avg       0.49      0.49      0.49    322950
weighted avg       0.50      0.51      0.50    322950
```

Fig. 10, Severity Rating Classification Report

It should be pointed out that if a particular classification model trained in one country or region is applied in another country or region, the prediction algorithm will then need to be modified accordingly. That is, the classification model needs to be re-trained based on a historical accident database from the country or region concerned.

**Common-sense Inferencing Engine (CIE)**. The ERE prediction will also output the corresponding pre-crash event of a certain accident as well, that is, the trigger event of the accident. The human brain can use common sense to speculate how the current scene will develop. The goal of using CIE is to infer whether the current scene will mature into the trigger of a type of accident. To this end, CIE needs to have the capability of scripting an event chain.

Reference to case 5 in 3.3:

*If a football suddenly flies into the lane, it will be very likely followed by a chasing boy rushing into the lane.*

It is very easy for a human driver to infer that kids chasing the ball may very likely being occluded by vehicles parked on the roadside. We want the CIE of an ADSB to draw a similar conclusion by exercising common-sense inferring so that an early avoidance precaution can be applied.

The ontology architecture itself has certain reasoning capabilities, such as Progue and Ontological Modelling Language (OML) languages. Common-sense knowledge bases such as ConceptNet and WordNet also have common-sense deduction capabilities. However, ConceptNet is oriented to objects and their features, while WordNet pays more attention to rhetorical associations. ATOMIC [20] pays more attention to event description, so it is more desirable to use ATOMIC as a common-sense reasoning engine.

Within the scope of existing common-sense knowledge, ATOMIC can output relevant event chain inferences responding to the inquiry inputs in the traffic safety related fields. Take the above ball-child case as an example; test runs results are shown in Table 7. In the inference results, "a car hits a person" is exactly what we wanted from CIE.

Table 7 - ATOMIC Inferences Results

| Test-run Inputs | ATOMIC Relations Outputs | ATOMIC Inference Returns |
|---|---|---|
| Test-run #1: "a ball is rolling at the intersection" | Happens Before | a football is thrown |
| | Happens After | a car hits the football |
| | | a car hits a person |
| Test-run #2: "x is chasing a ball" | X want | to catch the ball |
| | | to win the game |
| | | to get the ball |
| | X need | to have a ball |
| | | to go outside |
| | Happens Before | person X throws the ball to person Y |
| | | person X catches the ball and throws it |
| | | person X catches the ball |
| | Happens After | person X sees a ball on the ground |
| | | person X sees person Y throwing the ball |

Furthermore, we can handcraft more common-sense knowledge specific to the driving safety domain by defining relationships between elements. Following an example given in [20], a local ontology makes a point of the case "a vehicle entering the 'Exit Only' lane may suddenly return to the main lane and cut in in front of you" in 3.3 may be defined as in Fig. 11.

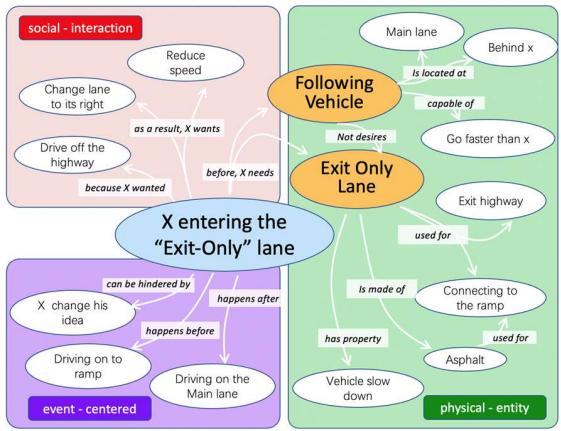

Fig. 11, "Drive-off" Case Related Relation Inputs

We ran Test-run #1 on Ubuntu 22.04 with processor of 2.4 GHz Core i5 and a memory of 16GB; it took about 5~6s to complete the inference computing on the ATOMIC common-sense knowledge-trained model (COMET-BART model). That is too long for a real-time ADSB application. To lightweight the CIE for the purpose of facilitating the on-board application, a lightweight common-sense knowledge base dedicated to traffic safety needs to be applied.

**Scene Perception System.** Unlike the tactical brain, the ADSB is focused on perceiving the environmental world at the scene level, that is, it needs to qualitatively perceive the classes of objects, the states of actors, and the associative properties between elements. In contrast, the tactical brain is focused on quantitative measurements of the physical properties of objects. Consequently, scenes must be decomposed, represented, perceived and modelled in the first place. In the ADSB model, an accident is a scene represented by element attributes. This requires the perception system to be equipped with the minimum basic capabilities in order to sense—in whatever way—81 elements involved in the ERE training. Moreover, the ADSB perception system should also be able to sense 44 pre-crash scenes typologized by NHTSA in [38]. Possible sensing channels are listed in Table 8.

Table 8 - Scenario Sensing Channels

| Sensing Channel | Elements |
|---|---|
| GPS receiver, digital map | State, county, city, mile point, latitude, longitude |
| digital clock | month of crash, day of crash, day of week, year of crash, hour of crash, minute of crash |
| GPS receiver, digital map | Relation to junction, trafficway identifier, route signing, roadway function class, type of intersection, trafficway description, |
| GPS receiver, digital map, highly autonomous driving (had) map | Total lanes in roadway, speed limit, roadway alignment, roadway grade, |
| On-board sensing | Roadway surface type, roadway surface condition, light condition, travel speed, |
| On-board sensing, network services | Atmospheric conditions |
| HMI (human-machine-interface) input by driver and automated driving controller developer | Body type, vehicle trailing, gross vehicle weight rating, vehicle configuration, cargo body type, bus use, special use, driver's zip code, age, sex |

**Goal and Value Setting.** Low-level intelligent machines exhibit two opposite extreme characteristics, either being aimless and emotionless, or sticking to a dead end and going all the way. This is the main reason why tactical-level AD systems cannot adapt to changes in the environment. The human way of thinking is often characterized as either logical or rational, but in reality, the presence of human emotions gives rise to multiple ways of thinking with many shades between the two extremes; [11] lists 19 of the typical approaches (pp 226). The human brain needs to switch between these approaches because it has a clear task goal in mind before that. This section discusses how intelligent machines can be goal oriented.

Goals and values give machines emotion which enables an intelligent machine to be resourceful in solving situation problems. A safe state of driving is one of the goals and values that AD systems can have. Safe states are classified into empirically safe state, theoretically safe state, and rule-of-thumb safe state here in this paper. There might be other goals and values in terms of driving comfort and agility.

(a) Empirically Safe State

If an AD vehicle drives in a scene that triggers no warning from the ERE whatsoever, we consider this AD vehicle is driving in an empirically safe state. RER can prevent an AD vehicle from driving in dangerous state experienced by FARS/CRSS records.

In addition to the warnings issued by the trained experience prediction model, another example of using experiences could be, if a heavy rainfall is forecast, to re-route the AD vehicle's driving plan in order to keep away from any blackspots of recorded flooding areas.

(b) Theoretically Safe State

Driving in a theoretically safe state means a subject vehicle keeping proper relative distances from other objects both longitudinally and laterally, choosing the most appropriate driving lane, and maintaining a reasonable surrounding object pattern in order to maintain a safe cruise under normal (ordinary) driving conditions. The safety criteria are formulated by domain experts.

For example, let $L_f$, $L_r$ and $L_t$ be the distances of the subject vehicle with respect to a leading vehicle, a trailing vehicle, and a lateral neighbouring vehicle. The subject vehicle is travelling at a given longitudinal speed $v_s$, and has a maximum accelerating capacity of $+a_{max,s}$ and a maximum braking capacity of $-a_{max,s}$. The leading vehicle is travelling at the given longitudinal speed $v_f$, and has a maximum accelerating capacity of $+a_{max,f}$ and a maximum braking capacity of $-a_{max,f}$. The trailing vehicle is travelling at the given longitudinal speed $v_t$, and has a maximum accelerating capacity of $+a_{max,t}$ and a maximum braking capacity of $-a_{max,t}$. The safety state of the subject vehicle could be defined by five aspects:

$L_f = a_r \cdot a_w \cdot a_l \cdot f(v_f, v_s, \pm a_{max,f}, \pm a_{max,s})$

$L_t = a_r \cdot a_w \cdot a_l \cdot f(v_r, v_s, \pm a_{max,t}, \pm a_{max,s})$

$L_l = a_r \cdot a_w \cdot a_l \cdot f(v^n_l, v^s_l, a^n_{max,l}, a^s_{max,l})$

*Proper lane use = 1, 2, 3....n*

*Maintaining a safety surrounding pattern*

where $a_r$, $a_w$ and $a_l$ are correction factors considering the road condition, weather condition and light condition at the present time, and superscripts $n$ and $s$ denote the lateral neighbouring vehicle and the subject vehicle respectively.

Function $f$ embodies the technical specifications widely used in automated vehicle engineering and the common-sense knowledge doctrines instructed by defensive driving practices, by considering the consequences resulting from vehicle kinetic and dynamic factors.

For example, a minimum longitudinal and lateral distance principal RSS (responsibility-sensitive safety) proposed by Mobileye [45] could be applied here.

When the ADSB system detects a driving situation that is deviating from a normal condition, coefficients are applied to the safety criteria. Taking trailing distance control for instance, the widely accepted rule requires that a driver should ideally stay at least two seconds behind any vehicle that is directly in front of his or her vehicle. But when adverse driving conditions arise that require a longer response time, such as driving on a slippery road surface in dark lighting conditions, the two-second rule may be modified to a two-and-a-half-second or three-second rule. These baseline rules are input by the transportation domain experts based on best practices recognised in the automobile industry. When a driving status deviation from the safety rules is detected, the ADSB will issue a warning to bring the vehicle back to a safe cruise status.

Safety surrounding pattern includes scenery aspect and actor aspect. Safety scenery attributes can be found by ERE from accident databases. We use a position coding system to describe the topological relationships relative to the surrounding actors, as shown in Fig. 12.

| -LL2 | -LL1 | LL | LL1 | LL2 |
|------|------|----|----|-----|
| -L2  | -L1  | L  | L1 | L2  |
| -2   | -1   | Subject Vehicle | 1 | 2 |
| -R2  | -R1  | R  | R1 | R2  |
| -RR2 | -RR1 | RR | RR1 | RR2 |

Forward Direction ⟶

Fig. 12, Position Coding for Topological Relationships

(c) Safe State of Rule of Thumb

A defensive driving course always suggests being alert for situations that box you in, and to adjust your speed so that you are not sandwiched between cars in the lanes on either side of you. A safety surrounding pattern can ensure the ego vehicle maintains an escape route in the case of a collision directly in front.

A "no zone" refers to the blind spots of a large vehicle—such as a bus or lorry—where one shouldn't be driving in, as shown in Fig. 13. A safety surrounding pattern can exclude a no-zone from the drivable space.

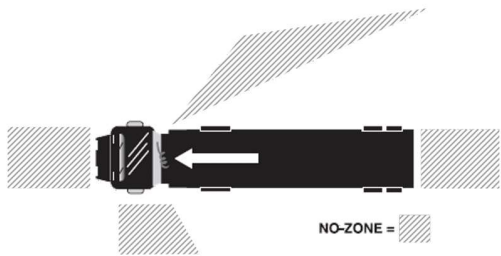

Fig. 13, "No Zone" of a Large Vehicle
(https://dmv.ny.gov)

A surrounding pattern can also be used by the CIS to infer which one actor of the neighbours can pose a potential threat.

Similar to the safe-state-keeping, an automated vehicle's behavior control can also be accomplished by coding generally accepted goals and values into the domain common-sense reservoir. An automated vehicle's normative behavior can give the occupants a feeling of riding smoothly and travelling swiftly, performing maneuvers with ample warning to other traffic participants, and helping the traffic to flow smoothly. Examples of such maneuvers and warnings include indicating an intended turn well in advance and braking smoothly over a longer period. The code of conduct for an automated vehicle driving in normal conditions can be defined as a baseline. When the driving environment deviates somewhat from the normal, the behavior criteria for a maneuver vary accordingly. For typical behavior elements subject to modification, see Table 9.

Rule-of-thumb and code-of-conduct entries can be edited manually according to the ontology protocol set out by 3.4.

Table 9 - Behavior Elements

| Event | Behavior Element that Needs Control |
|---|---|
| Start | Accelerating strategy |
| Stop | Braking strategy |
| Approach | Pull up deceleration; pull over deceleration; pull over trajectory, distancing |
| Overtaking | Lateral passing distance, longitudinal passing distance; passing speed |
| Cornering | Decelerating-accelerating profile |
| Curve negotiating | Speed |
| Interaction | Intention articulating |

Goals and values are hand-written by system designers and are performed by the Goal & Value Keeper (GVK) shown in Fig. 14.

**ADSB System application.** The Scene Perception System (SPS), ERE, CIE, and Goal and Value Keeper (GVK) work together to act as a strategical planning brain, as shown in Fig 14.

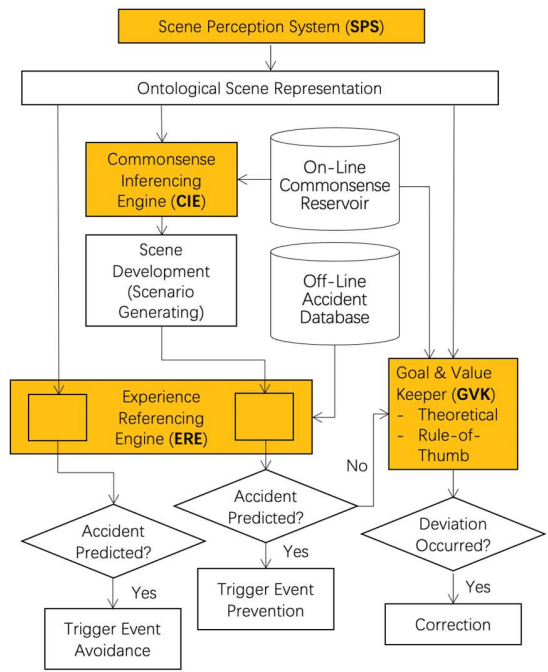

Fig. 14 ADSB Architecture Pillared by Three Kernel Engines

Figure 15 shows a block diagram of a combined application wherein an ADSB system is integrated with conventional motion controller architecture. Take safety state keeping for example, when the subject vehicle intends to perform a state transition from State A to State B for example, a mission planning instruction is first transmitted as a request to the scene sensing system, rather than being outputted immediately to the control unit. The request seeks a safety evaluation of State B. It is only when the ADSB considers State B to be safe that the strategical monitor outputs the instruction back to mission planning with an accompanying "GO" instruction. The mission planning then outputs the instruction to the control unit of the motion controller. If the ADSB considers State B to be unsafe, it will inhibit or cancel the instruction. This can happen if, for example, some potential hazards associated with the instruction are identified by the ADSB system.

Commonsense-augmented perception is another example of dual-brain application. If ADSB sees a slow-moving heavy machine working on the heavy snow-covered road, it might conclude that it is a snow removal truck with high probability by exercising commonsense inferencing, without having to be pre-trained with visuals.

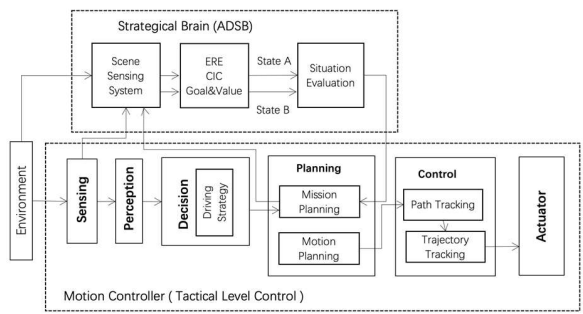

Fig. 15, ADSB Integrated with a Tactical Motion Controller

Figure 16 is a block diagram of an example application of the ADSB system, which is used as an after-market advance driver-assistance system (ADAS). The ADSB gives advice on matters such as desired driving speed, minimum peripheral distancing, and appropriate lane usage etc., to a human driver. Based on scenario information collected by a scene perception system, the ADSB system recommends safe speed, topological surrounding pattern, and lane positioning to a driver through the output of one or more HMI output devices, which can employ voice, screen display, vibration, and other multiple kinds of communication channels. Before starting to drive, the driver can input background information through one or more HMI input devices, which can take the form of touchscreen displays, voice input, and other communication channels. The background information typically includes vehicle information such as vehicle type, weight, body type, application (or use), etc., as well as driver status information such as gender, age, driver's licence issuance location, and so on.

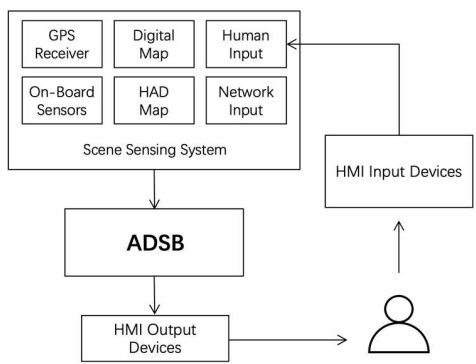

Fig. 16, A Block Diagram of an Example Application of the ADSB System

## 5. Conclusions

We present the concept of an Automated Driving Strategical Brain (ADSB): a framework of a scene perception and evaluation system that works at a higher abstraction level, incorporating experience referencing, common-sense inferring and goal-and-value judging capabilities, to provide a contextual perspective for decision making within AD planning.

ADSB can be used to provide directional advice when tactical-level environmental perception conclusions are ambiguous; it can also use future scenario models to remind tactical-level decision systems to plan ahead when a potential hazard scene is perceived. ADSB can also be installed in the aftermarket as a stand-alone system to provide drivers with fail-soft advice for safe driving.

An experience-referencing model was trained using the FARS/CRSS database. The prediction accuracy for occurrence probability and type of crash of severe accident is encouragingly good. The prediction accuracy for severity classification is acceptable, considering this early development stage.

A trained model, COMET-BART by ATOMIC, [46] was used as the kernel part of a common-sense inference engine (CIE). Test runs yielded the desired relevant event inference, nevertheless, the time it took was exceedingly long for the purpose of on-board application. As a result, a road-traffic-specific-domain, common-sense, lightweight knowledge base is needed for these ends.

We believe that with good scalability, the ADSB approach provides a potential solution to the problem of long-tail corner cases encountered in the validation of a rule-based planning algorithm. In fact, element-based scenes and goals can be further decomposed into subscenes and subgoals. A difference engine is used by [11] to generate such goals. Solution synthesising for unfamiliar driving environments could be realised by combining multi-solutions for multi-subgoals in further research.

A scene perception system capable of sensing all the input information set out by Table 8 remains to be developed.

Verification and validation guidelines for evaluating the effectiveness of the ADSB remain to be conceived.

The authors believe that a human-like dual-brain planning architecture will prevail in the future AD vehicle industry to compensate for the defects and insufficiencies of current cognitive technologies. ADSB will facilitate many exciting research fields in this regard.

-------------------------------------------------------------------